%% file: main.tex
\documentclass[10pt,twocolumn,letterpaper]{article}

\usepackage[pagenumbers]{cvpr} 

\usepackage{pifont}
\usepackage[dvipsnames]{xcolor}

\definecolor{cvprblue}{rgb}{0.21,0.49,0.74}
\usepackage[pagebackref,breaklinks,colorlinks,allcolors=cvprblue]{hyperref}

\def\paperID{HOW-32} 
\def\confName{CVPR}
\def\confYear{2026}

\usepackage{tikz}

\newcommand\copyrighttext{%
  \footnotesize \textcopyright \the\year{} IEEE. Personal use of this material is permitted. Permission from IEEE must be obtained for all other uses, in any current or future media, including reprinting/republishing this material for advertising or promotional purposes, creating new collective works, for resale or redistribution to servers or lists, or reuse of any copyrighted component of this work in other works. IEEE/CVF Conference on Computer Vision and Pattern Recognition (CVPR) Workshops.}

\newcommand\copyrightnotice{%
\begin{tikzpicture}[remember picture,overlay]
\node[anchor=south,yshift=2pt] at (current page.south) {\fbox{\parbox{\dimexpr0.75\textwidth-\fboxsep-\fboxrule\relax}{\copyrighttext}}};
\end{tikzpicture}%
}

\title{Differences in Detection: Explainability Where it Matters}

\author{Johannes Theodoridis\\
University of Tübingen\\
Institute for Applied AI\\
{\tt\footnotesize Jo.Theodoridis@googlemail.com}
\and
Johannes Maucher\\
Hochschule der Medien Stuttgart\\
Institute for Applied AI\\
{\tt\footnotesize maucher@hdm-stuttgart.de}
\and
Andreas Schilling\\
University of Tübingen\\
{\tt\footnotesize schilling@uni-tuebingen.de}
}

\begin{document}
\maketitle
\copyrightnotice

\input{sec/0_abstract}    
\input{sec/1_intro}
\input{sec/2_dnd}
\input{sec/3_limitations}
\input{sec/4_conclusion}
\clearpage
{
    \small
    \bibliographystyle{ieeenat_fullname}
    \bibliography{dnd}
}


\end{document}

%% file: sec/0_abstract.tex
\begin{abstract}
We propose Differences in Detection (DnD), an intuitive method to compare two object detection models. Based on the same matching algorithm, it complements the standard metrics of mean Average Precision ($mAP$) and TIDE error analysis with the ability to compare two models directly. More specifically, we calculate the intersection of ground truth labels that are recognized by both models, followed by the corresponding difference sets and the complement set of ground truth labels that are missed by both models. The resulting comparison is more direct and intuitive than a comparison of independent summary statistics. It reveals individual and shared mistakes and becomes particularly interesting when combined with error types. In this case, the differences in detection errors can be analyzed naturally in a standard confusion matrix. While valuable in itself, we believe that one of the best applications of DnD is to guide explainability methods such as ODAM towards metric-relevant examples, grounded in structured subsets. The code for our method is available here:\\

\raggedright\noindent\href{https://github.com/JohannesTheo/differences-in-detection}{https://github.com/JohannesTheo/differences-in-detection}

\end{abstract}

%% file: sec/1_intro.tex
\section{Introduction}
\label{sec:intro}

\begin{figure}[t!]
  \centering
   \includegraphics[width=0.82\linewidth]{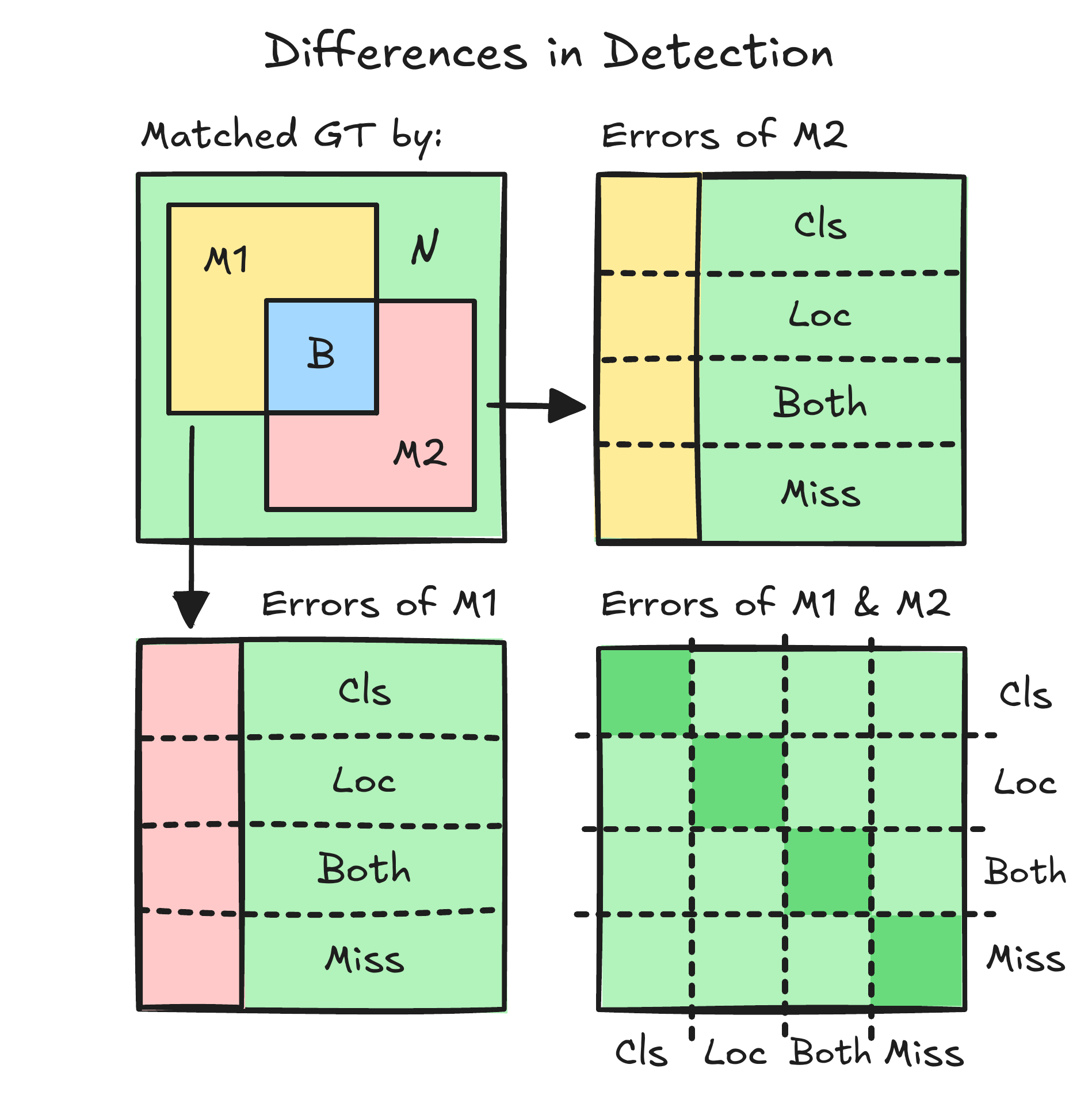}
   \caption{For the predictions of two models we calculate the intersection and difference of their matched ground truth labels. Unmatched instances can be labeled by the matching TIDE error. Together, this enables a structured comparison of predictions at the instance level which is not possible with standard $mAP$ or TIDE.}
   \label{fig:dnd}
\end{figure}

Analyzing object detection models can be easy and difficult at the same time. Standardized summary metrics such as mean Average Precision ($mAP$) enable a quick and concise comparison of overall performance and come with easily accessible implementations, e.g. \texttt{pycocotools} \cite{lin_microsoft_2014} or \texttt{faster-coco-eval} \cite{mixaill76_faster-coco-eval_2024}. At the same time, they can be difficult to interpret because localization requires us to select a specific Intersection over Union (IoU) threshold in order to calculate the precision-recall curve for detections. Since predictions must also be ranked based on their score, the resulting matching algorithm, but more importantly the final value of the IoU- and class-averaged metric, is not particularly intuitive. If a model gains $+0.5$ $mAP$ on a specific benchmark, for instance on MS-COCO \cite{lin_microsoft_2014}, what does that mean? With a little effort, we can also compare the mean Average Recall ($mAR$) or the class-wise $AP$ values of two models, but more insight becomes challenging. For analyzing the sources of errors, \citet{bolya_tide_2020} introduced the TIDE toolbox. It defines six types of error and measures the expected gain in $mAP$ when errors are fixed \textit{individually}. Previous methods calculated errors \textit{progressively} \cite{lin_microsoft_2014,borji_empirical_2019}, conveniently adding up to $100$ $mAP$, but greatly overestimating the impact of the last corrected error type. Together, summary statistics and TIDE errors can provide a good overview, but still have a common limitation. In both cases, models are evaluated in isolation, and we do not know whether two models make similar or different errors, or whether they even recognize the same ground truth instances. At the other end of the spectrum, visual explainability methods for object detection \cite{selvaraju_grad-cam_2017, chattopadhay_grad-cam_2018, petsiuk_black-box_2021, yamauchi_spatial_2022, zhao_odam_2023, yamauchi_spatial_2024, xia_explaining_2025} enable true in-depth analysis at the prediction level. Methods like Object Detector Activation Maps (ODAM) \cite{zhao_odam_2023} produce heatmaps that can help us interpret model decisions. A key benefit of our method is that it provides natural and intuitive subsets to guide these methods. Consider the following example. The \texttt{val2017} split of MS-COCO has $5000$ images and $37k$ ground truth annotations. The re-annotated variants Sama-COCO \cite{zimmermann_benchmarking_2023} and COCO-ReM \cite{singh_benchmarking_2025} even have $41k$ and $47k$ annotations respectively. For a single image, a common evaluation protocol is to enforce a strict maximum of $100$ detections. Even though the theoretical limit of $500k$ is usually not reached, object detection models often exceed the $100k$ mark in practice, which is still far too many predictions to search and analyze manually.\\

In the following, we present a simple solution to both problems. It complements the strengths of $mAP$ and TIDE errors with the ability to compare two models directly and provides structured subsets of relevant predictions that can be used to apply explainability methods where it matters.

%% file: sec/2_dnd.tex
\section{Differences in Detection}
\label{sec:dnd}

We propose a direct comparison of object detection models by analyzing their differences in detection on a complete dataset. More precisely, we are interested in the similarities and differences between the matching and missing ground truth labels. Assuming two models $M1$ and $M2$, we first apply the same matching algorithm as $mAP$ to the respective prediction sets $D1$ and $D2$. Similarly to $mAP$, predictions can be anything that supports an IoU-based definition of true positives such as bounding boxes, instance masks, or object boundaries \cite{cheng_boundary_2021}. However, instead of calculating summary metrics, we extract the \textit{matching pairs} of predictions and ground truth labels $(dt_{i}, gt_{m})$ and $(dt_{j}, gt_{m})$ with $dt_{i} \in D1$ and $dt_{j} \in D2$. Since both models are linked through the shared set of \textit{ground truth labels} $GT$, we can now calculate more interesting subsets as displayed visually in \Cref{fig:dnd}. For simplicity, we will reuse the model notation and refer to these subsets as follows:

\begin{align}
    &                      & & \text{Matched GT by:} \nonumber  \\
  B & = D1 \cap D2         & & \text{both}           \label{eq:A} \\ 
  M1 & = D1 - B            & & \text{only } D1       \label{eq:B} \\ 
  M2 & = D2 - B            & & \text{only } D2       \label{eq:C} \\ 
  N & = GT - (D1 \cup D2)  & & \text{neither}        \label{eq:D} \\ 
    &                      & & \nonumber \\
    &                      & & \text{Unmatched GT by:} \nonumber \\
E1 & = N \cup M2 & & D1 \label{eq:EB} \\ 
E2 & = N \cup M1 & & D2 \label{eq:EC}
\end{align}

In contrast to summary statistics like $mAP$, our direct comparison provides intuitive and natural subsets which can be helpful to select examples for in-depth analysis. As an alternative, we could also match $D1$ and $D2$ directly, but this would not guarantee that the resulting pairs would be relevant to the metric we are ultimately interested in. Within our approach, the error sets $E1$ and $E2$ are particularly interesting since they include the positive predictions of the other model $M2$ and $M1$, respectively. By definition, the latter are equivalent to the corresponding difference sets that contain the individual errors. Together, they allow us to analyze paired examples of failure and success, which is intriguing.
\begin{align}
    &                     & & \text{Individual Errors of:} \nonumber     \\
E\mathrm{x}1 & = E1 - N  = M2 & & M1 \label{eq:EB} \\ 
E\mathrm{x}2 & = E2 - N  = M1 & & M2 \label{eq:EC}
\end{align}

So far, unmatched $gt$ instances are defined as \textit{missed} and have only one $dt$ relation for individual errors. Since we use the same matching algorithm as $mAP$, we can reuse the error correction of TIDE and increase the depth of our analysis. For every unmatched $gt_u$, we query the error oracle as shown in \cref{tab:tide}, and extract the respective matching triplet $(e, dt_i, gt_u)$ and $(e, dt_j, gt_u)$ that would fix it.

\begin{table}[h!]
  \caption{TIDE Error Analysis: Candidate $dt$ are sorted by score before matching. By default, the background and foreground IoU thresholds $t_b$ and $t_f$ that define false and true positives are set to $0.1$ and $0.5$. Please see \citet{bolya_tide_2020} for a complete explanation.}
  \label{tab:tide}
  \centering
\begin{tabular}{l c c c}
    Error & Class & IoU & Oracle \\
    \midrule
    \texttt{Cls}  & \textcolor{red}{\ding{55}}         & $> t_f$ & \textbf{match} $\boldsymbol{gt}$\\
    \texttt{Loc}  & \textcolor{ForestGreen}{\ding{51}} & $< t_f$ & \textbf{match} $\boldsymbol{gt}$ \\
    \texttt{Both} & \textcolor{red}{\ding{55}}         & $< t_f$ & ignore $dt$ \\
    \texttt{Dupe} & \textcolor{ForestGreen}{\ding{51}} & $> t_f$ & ignore $dt$ \\
    \texttt{Bkg}  & \textcolor{Apricot}{\ding{107}}    & $< t_b$ & ignore $dt$ \\
    \texttt{Miss} & --                                 & --      & \textbf{ignore} $\boldsymbol{gt}$
\end{tabular}
\end{table}

Importantly, we only consider false negatives of type \texttt{Cls}, \texttt{Loc} and \texttt{Miss}, which can be related to specific $gt$ instances. The remaining error types are false positives and are fixed by ignoring the respective $dt$. To gain a little more insight, we deviate from TIDE and implement a matching oracle for \texttt{Both} errors as well. With everything in place, we can now compare the distributions of individual errors $E\mathrm{x}1$ and $E\mathrm{x}2$ separately, and the shared errors $N$ in a confusion matrix. In both cases, we treat the assigned error types as categories.\\

In the following, we present two example applications of our proposed method in the context of MS-COCO. Specifically, we show how it complements a standard comparison with $mAP$ and TIDE. In general, our tool can be used with any dataset that complies with the \texttt{COCO API} format \cite{lin_microsoft_2014}.

\subsection{Performance comparison:}

\begin{figure}[h!]
  \centering
   \includegraphics[width=.825\linewidth]{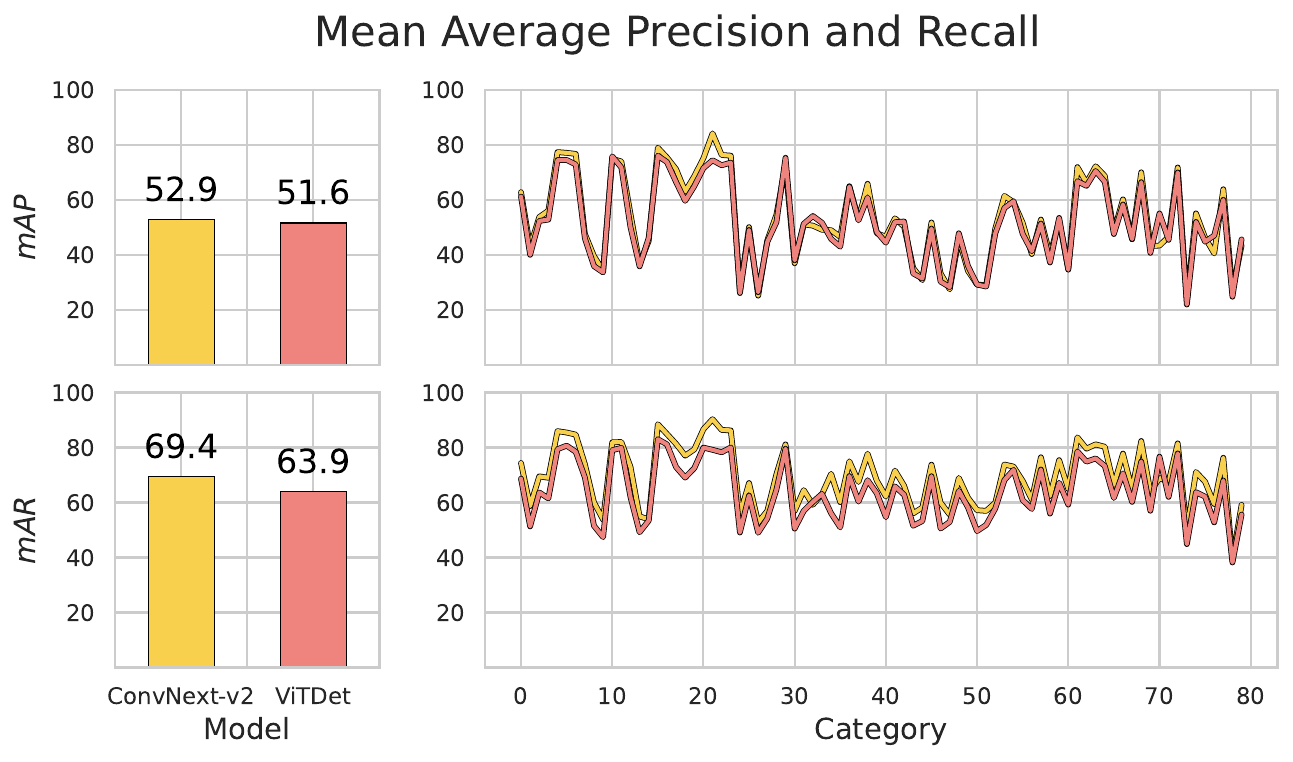} \\
   \vspace{8pt}\hspace{15pt}
   \includegraphics[width=.725\linewidth]{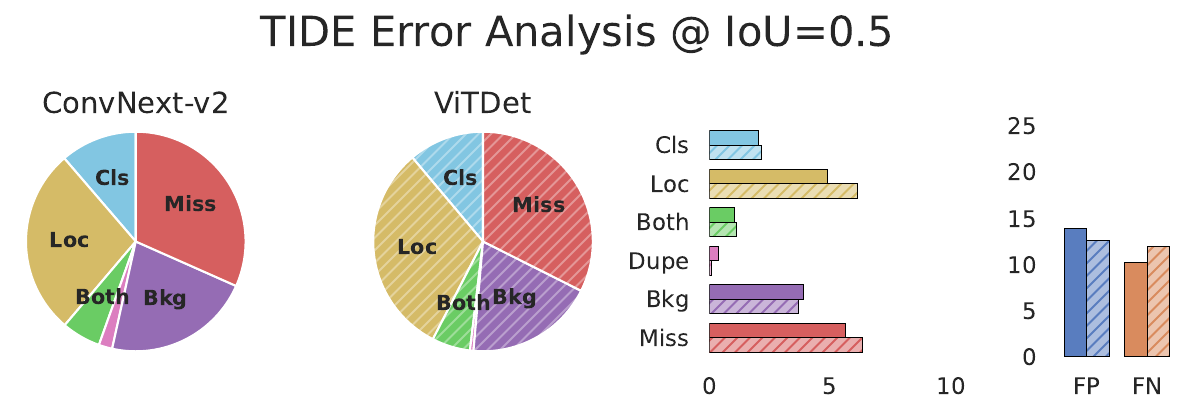}
   \caption{Comparison of summary metrics and TIDE errors for Mask R-CNN with ConvNext-v2-B and ViTDet-B on MS-COCO.}
   \label{fig:summary_1}
\end{figure}

In our first experiment, we compare Mask R-CNN \cite{he_mask_2017} with two backbone architectures on the popular MS-COCO benchmark. We choose ConvNext-v2-B \cite{woo_convnext_2023} and a plain VisionTransformer (ViT) \cite{dosovitskiy_image_2021} in the ViTDet-B \cite{li_exploring_2022} configuration as our models. Both are initialized from pretrained MS-COCO weights, trained with Large Scale Jitter (LSJ) \cite{ghiasi_simple_2021} data augmentation,
and both are evaluated on the \texttt{val2017} split. In \Cref{fig:summary_1}, we display the summary metrics $mAP$, $mAR$ and a standard TIDE error analysis for bounding box predictions. As can be seen, ConvNext-v2 has a slightly higher $mAP$ but a much better $mAR$ compared to ViTDet. This is also reflected in the TIDE error analysis, where ViTDet would benefit more from reducing false negatives, while ConvNext-v2 would benefit more from reducing false positive predictions. More insight can be gained from a per category comparison, but neither method allows us to analyze predictions at the instance level. For instance, do both models actually recognize the same objects, and, if not, do they make similar or different mistakes?

\begin{figure}[t!]
  \centering
   \includegraphics[width=.875\linewidth]{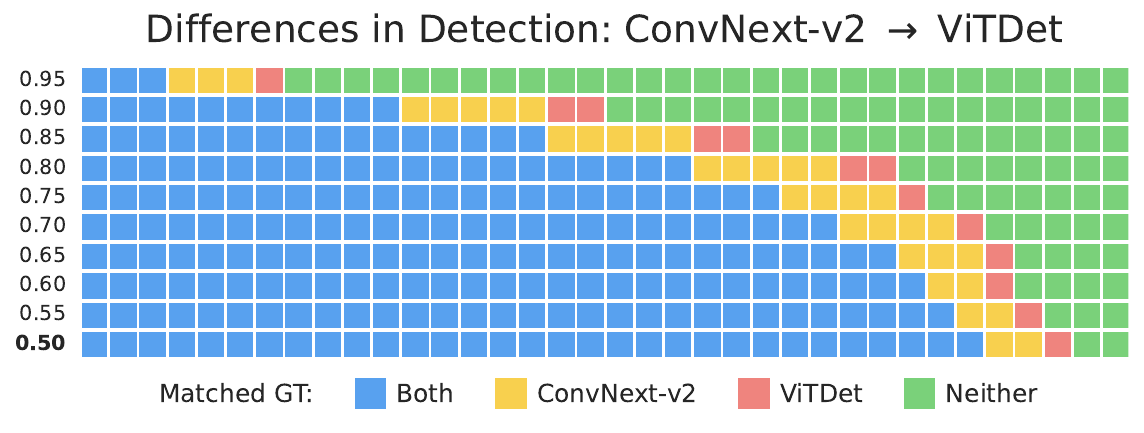}\\
   \vspace{8pt}
   \includegraphics[width=.875\linewidth]{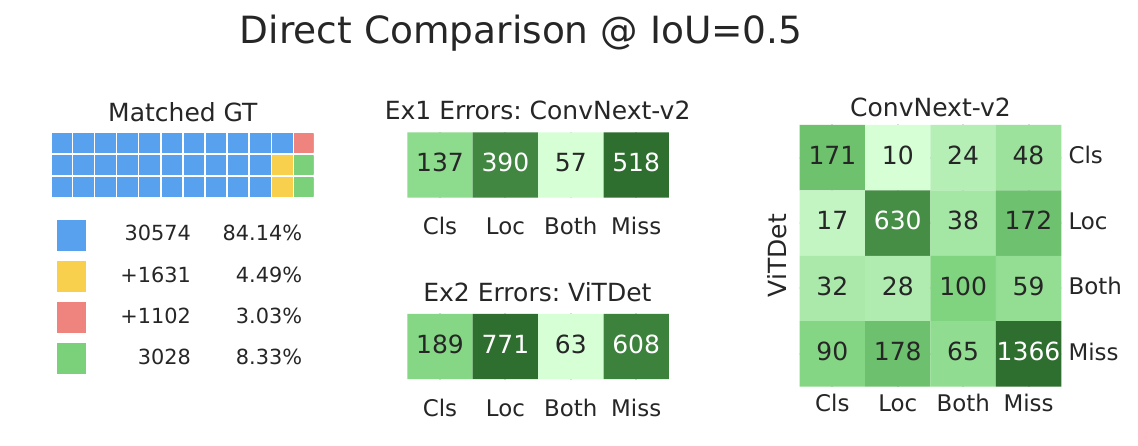}
   \caption{Differences in Detection for two Mask R-CNN models with the backbones ConvNext-v2-B and ViTDet-B on MS-COCO.}
   \label{fig:dnd_1}
\end{figure}

\begin{figure}[b!]
  \centering
   \includegraphics[width=.875\linewidth]{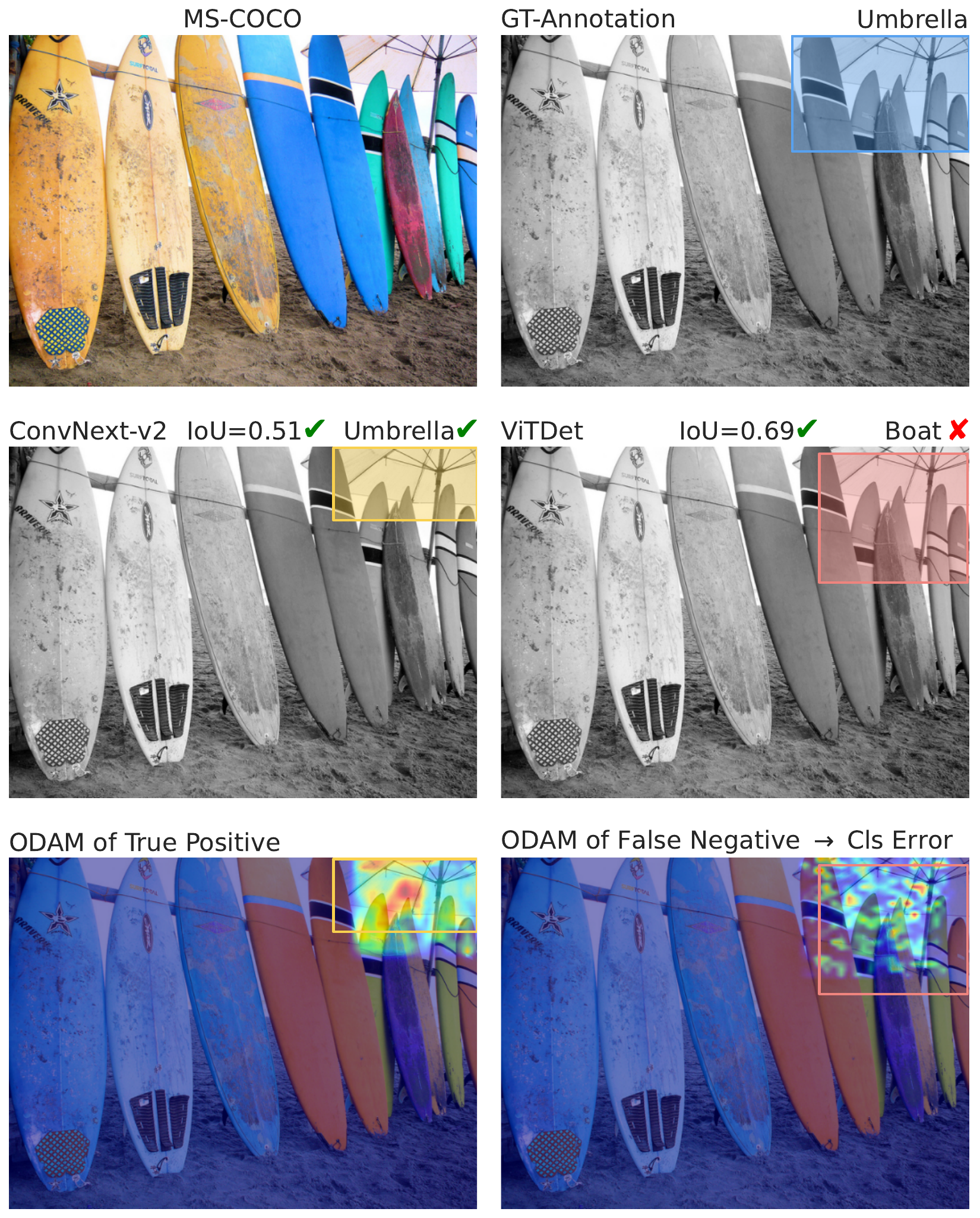}\\
   \caption{Example from the set of individual errors of ViTDet. ODAM visualizations display activations within proposal regions.}
   \label{fig:odam_1}
\end{figure}

As shown in \Cref{fig:dnd_1}, our proposed DnD method can easily answer these questions. Furthermore, it enables new and additional in-depth comparisons, which are not possible with $mAP$ and TIDE alone. For example, the shared sets of matched ($B$) and unmatched ($N$) instances can be used to uncover easy and hard examples in a detection dataset or to track previously matched instances throughout the training process. However, the core benefit of DnD is that it enables a structured and direct comparison of models at the instance level. In our case, it turns out that ViTDet actually detects instances $M2$ that ConvNext-v2 does not, despite its generally better performance and a much larger set of exclusive matches $M1$. The union of shared and individual detections, $B \cup M1$ and $B \cup M2$ is equivalent to the class-agnostic recall. Although interesting, we believe that DnD is used best as a starting point for explainability methods or for hypothesis testing. In the former case, the related sets of individual errors $E\mathrm{x}1$ and $E\mathrm{x}2$ provide paired examples for visualization as shown in \cref{fig:odam_1}. In the latter case, the confusion matrix of shared errors $N$ not only mirrors TIDE, but also holds references to ground truth annotations which can be used to precisely assess the effect of ablation studies.

\clearpage
\subsection{Robustness analysis:}

\begin{figure}[h!]
  \centering
   \includegraphics[width=.825\linewidth]{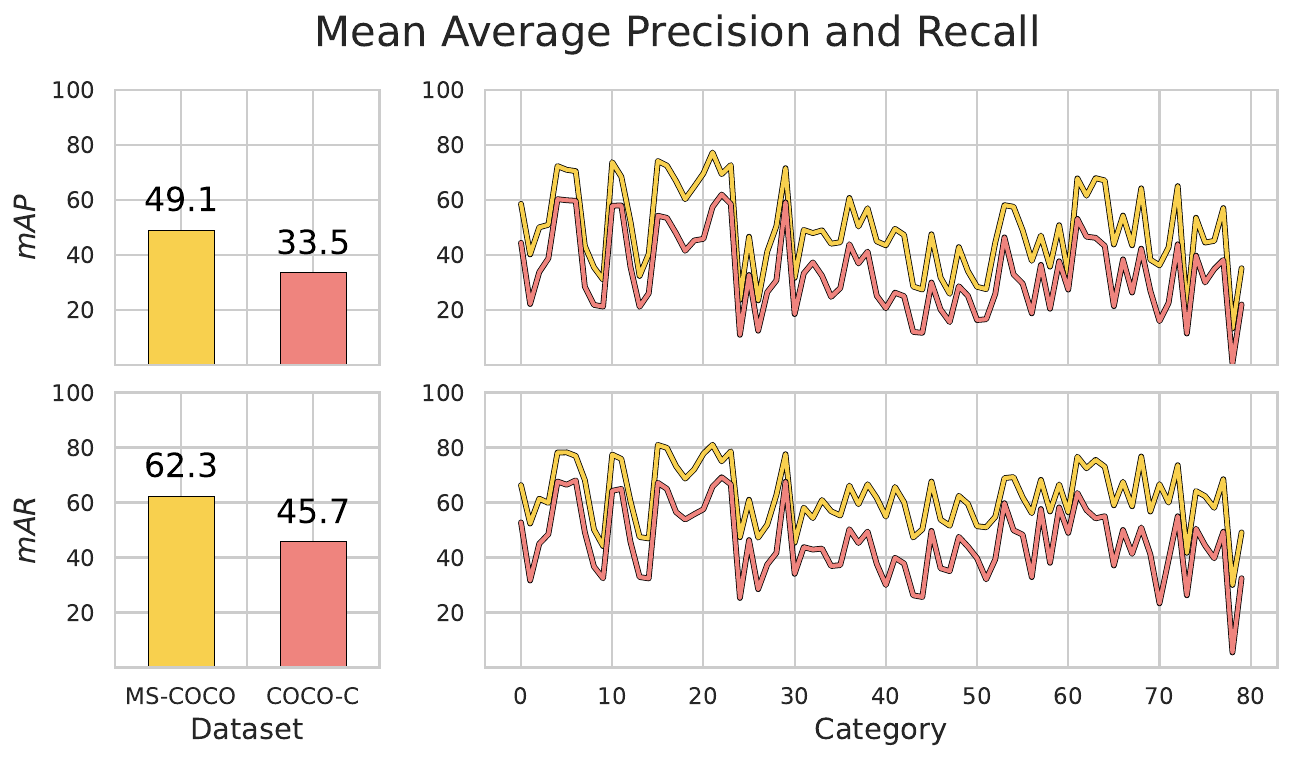} \\
   \vspace{8pt}\hspace{15pt}
   \includegraphics[width=.725\linewidth]{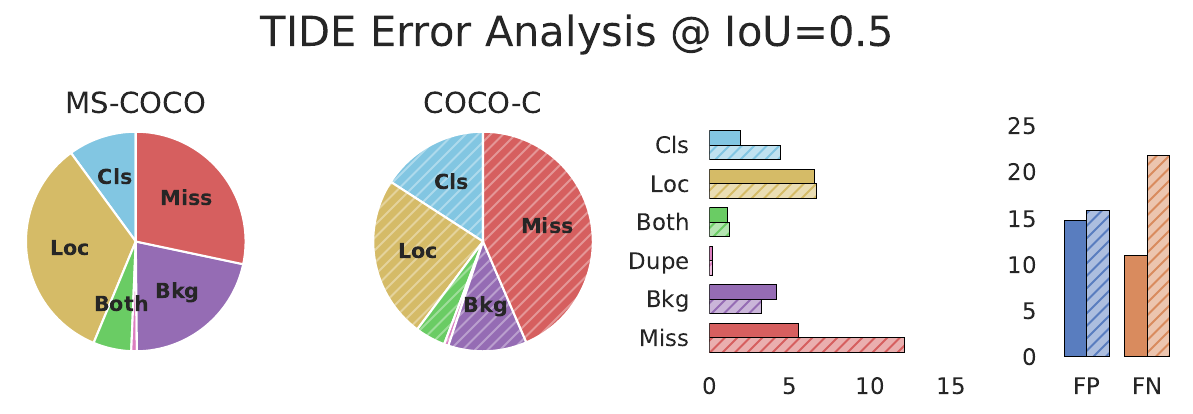}
   \caption{Comparison of summary metrics and TIDE errors for Mask R-CNN with FAN-S (hybrid) on MS-COCO and COCO-C.}
   \label{fig:summary_2}
\end{figure}

\begin{figure}[h!]
  \centering
   \includegraphics[width=.875\linewidth]{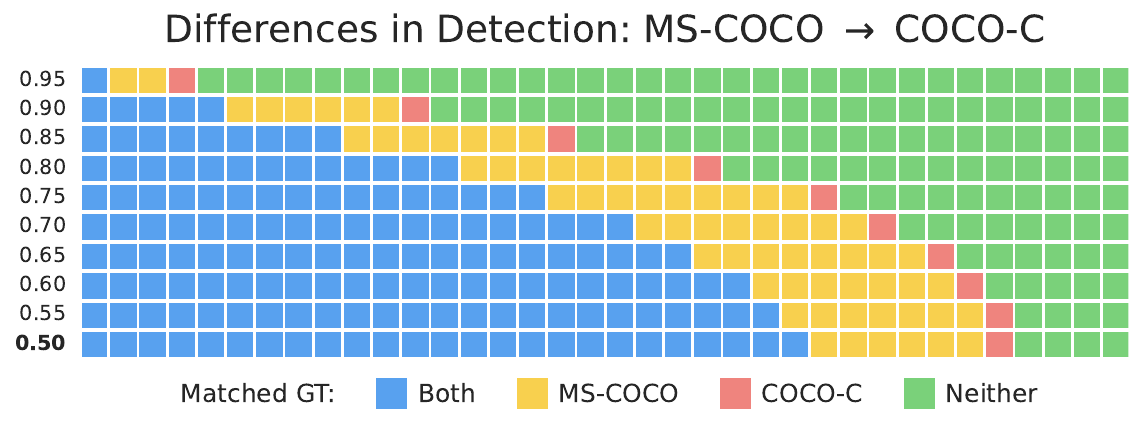}\\
   \vspace{8pt}
   \includegraphics[width=.875\linewidth]{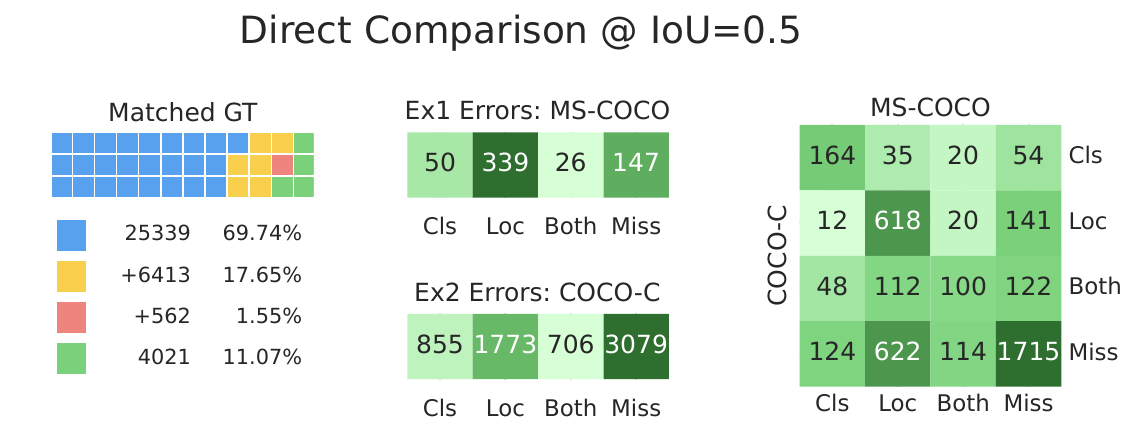}
   \caption{Differences in Detection for Mask R-CNN with FAN-S (hybrid) backbone evaluated on MS-COCO and COCO-C.}
   \label{fig:dnd_2}
\end{figure}

In our second experiment, we show an alternative use case of DnD and compare one model on two datasets. As our model, we select Mask R-CNN with a hybrid FAN \cite{zhou_understanding_2022} backbone that was developed to improve robustness. In this case, $D1$ now represents the predictions made on MS-COCO and $D2$ the predictions made on COCO-C \cite{michaelis_benchmarking_2019}, a challenging robustness benchmark. In general, other datasets with the same $GT$ annotations such as (Object-Centric) Stylized COCO \cite{michaelis_benchmarking_2019, theodoridis_trapped_2022} can be used as well. For our experiment, we choose the common corruption \texttt{Gaussian Noise} as proposed by \citet{hendrycks_benchmarking_2019} at medium severity level $3$. In \cref{fig:summary_2} we compare summary metrics and TIDE errors on MS-COCO and COCO-C and show the corresponding DnD in \cref{fig:dnd_2}.

\begin{figure}[h!]
  \centering
   \includegraphics[width=.875\linewidth]{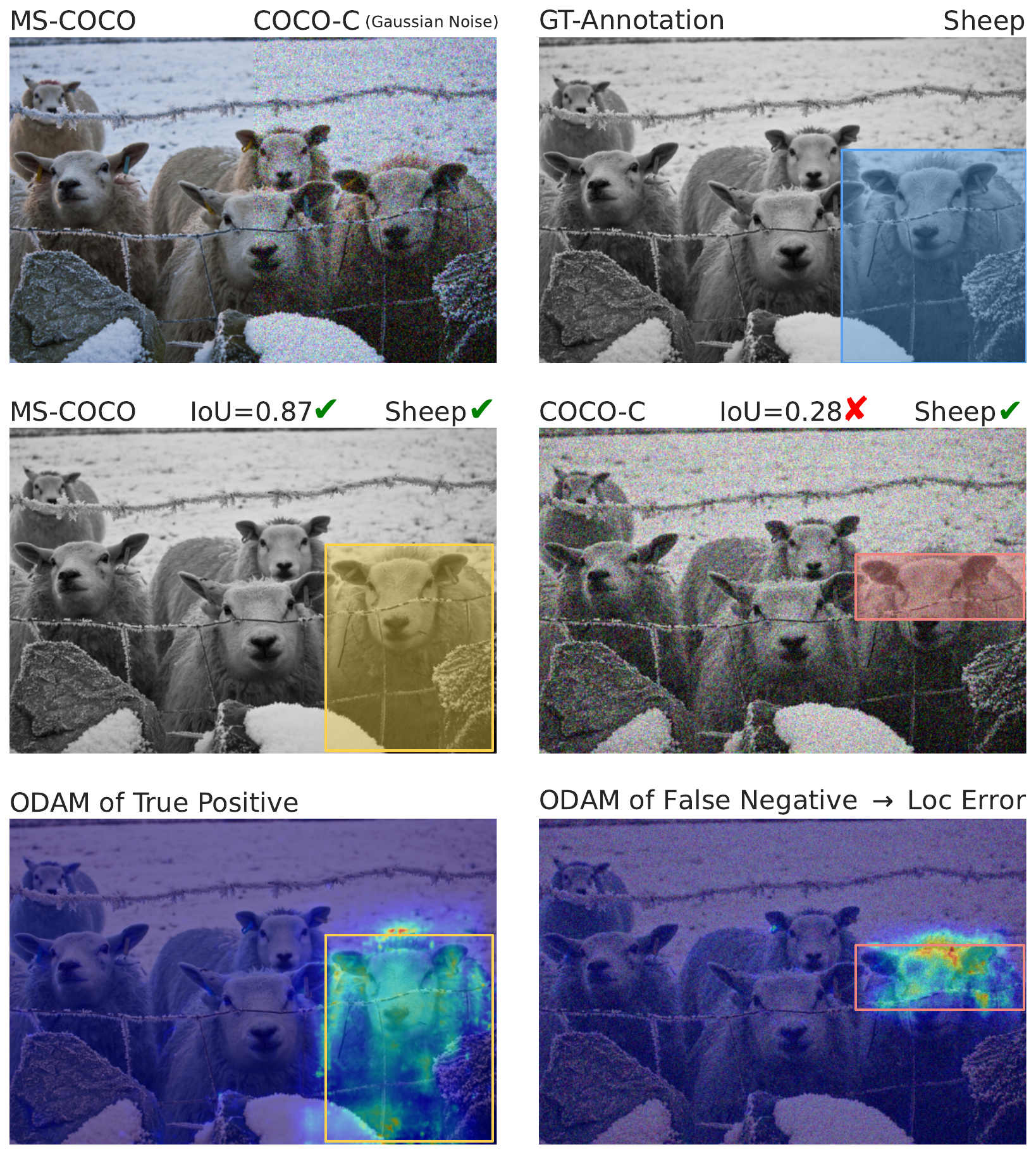}\\
   \caption{Example from the set of individual errors on COCO-C. ODAM visualizations display activations within proposal regions.}
   \label{fig:odam_2}
\end{figure}

As can be seen, Mask R-CNN with FAN-S is still very much affected by the added gaussian noise in COCO-C and we visualize one such example from the $E\mathrm{x}2$ set in \cref{fig:odam_2}. Interestingly, some previously unmatched instances are detected after adding noise, which is not intuitive or expected.

%% file: sec/3_limitations.tex
\section{Limitations}
\label{sec:limitations}

For more than two models, DnD must be applied in a chain of comparisons in its current form. However, extending the set definitions to the multi model case could be interesting, for instance to analyze the contributions of individual models in ensemble approaches. A second limitation is that DnD is defined from the $GT$ perspective. In consequence, it represents the class-agnostic recall and only considers positively matched predictions or TIDE errors that can fix a false negative $gt_u$ example. Extending DnD to false positive predictions as well would enable a direct comparison that is fully related to the $mAP$ metric, but requires to match the detections of two models directly, which is not well-defined and left as future work.

%% file: sec/4_conclusion.tex
\section{Conclusion}\label{sec:conclusion}

In this work, we propose Differences in Detection (DnD), an intuitive method to compare the predictions and errors of two object detection models directly, complementing the strength of $mAP$ and TIDE error analysis. Derived from the same matching algorithm, DnD subsets are well-defined and structured which is interesting in general but particularly useful for selecting metric-relevant examples for \mbox{explainability} methods in situations where it matters.